\documentclass[10pt,twocolumn,letterpaper]{article}

\usepackage[pagenumbers]{cvpr} 
\usepackage{algorithm}
\usepackage{comment}
\usepackage{amsmath}
\usepackage{amsfonts}
\usepackage[noend]{algpseudocode}

\usepackage{lipsum}
\usepackage{subcaption} 

\newcommand{\mm}[1]{\textcolor{blue}{MM: #1}}

\usepackage{array} 

\newcommand{\uncommon}{ uncommon }

\definecolor{cvprblue}{rgb}{0.21,0.49,0.74}
\usepackage[pagebackref,breaklinks,colorlinks,allcolors=cvprblue]{hyperref}

\def\paperID{28} 
\def\confName{CVPR}
\def\confYear{2025}

\title{Not all Views are Created Equal: \\ Analyzing Viewpoint Instabilities in Vision Foundation Models}

\author{Mateusz Michalkiewicz$^1$, Sheena Bai$^1$, Mahsa Baktashmotlagh$^2$, Varun Jampani$^3$, Guha Balakrishnan$^1$
\and
\\ 
$^1$ Rice University \\
$^2$ The University of Queensland \\
$^3$ Stability AI \\
}

\begin{document}

\maketitle

\begin{abstract}
In this paper, we analyze the viewpoint stability of foundational models - specifically, their sensitivity to changes in viewpoint- and define instability as significant feature variations resulting from minor changes in viewing angle, leading to generalization gaps in 3D reasoning tasks. We investigate nine foundational models, focusing on their responses to viewpoint changes, including the often-overlooked accidental viewpoints where specific camera orientations obscure an object's true 3D structure. Our methodology enables recognizing and classifying out-of-distribution (OOD), accidental, and stable viewpoints using feature representations alone, without accessing the actual images. 
 Our findings indicate that while foundation models consistently encode accidental viewpoints, they vary in their interpretation of OOD viewpoints due to inherent biases, at times leading to object misclassifications based on geometric resemblance. 
 Through quantitative and qualitative evaluations on three downstream tasks—classification, VQA, and 3D reconstruction—we illustrate the impact of viewpoint instability and underscore the importance of feature robustness across diverse viewing conditions.
\end{abstract}

    
\section{Introduction}
\label{sec:intro}

A common assumption in visual reasoning is that scene content in the 3D world is independent of camera viewpoint. While implicitly assumed in classical reconstruction and rendering algorithms~\cite{classic_recon}, this assumption is not explicitly enforced in feature representation models now used in a range of computer vision applications. Popular foundation representation models, e.g., DINO~\cite{dino} and CLIP~\cite{clip}, simply leverage large datasets and self-supervised training objectives to transform image pixels into a different feature space amenable to semantic interpretation. Although effective, this approach can introduce generalization gaps, particularly related to 3D reasoning~\cite{el2024probing,3d_gap} and spatial context~\cite{wang2024sam}, resulting in poor performance when faced with unusual viewpoints or atypical test cases. Hence, while the success of foundation models are impressive, they lack certain fundamental reasoning capabilities, such as content-viewpoint disentanglement. 

\begin{figure}[t!]
    \centering
    \includegraphics[width=\linewidth]{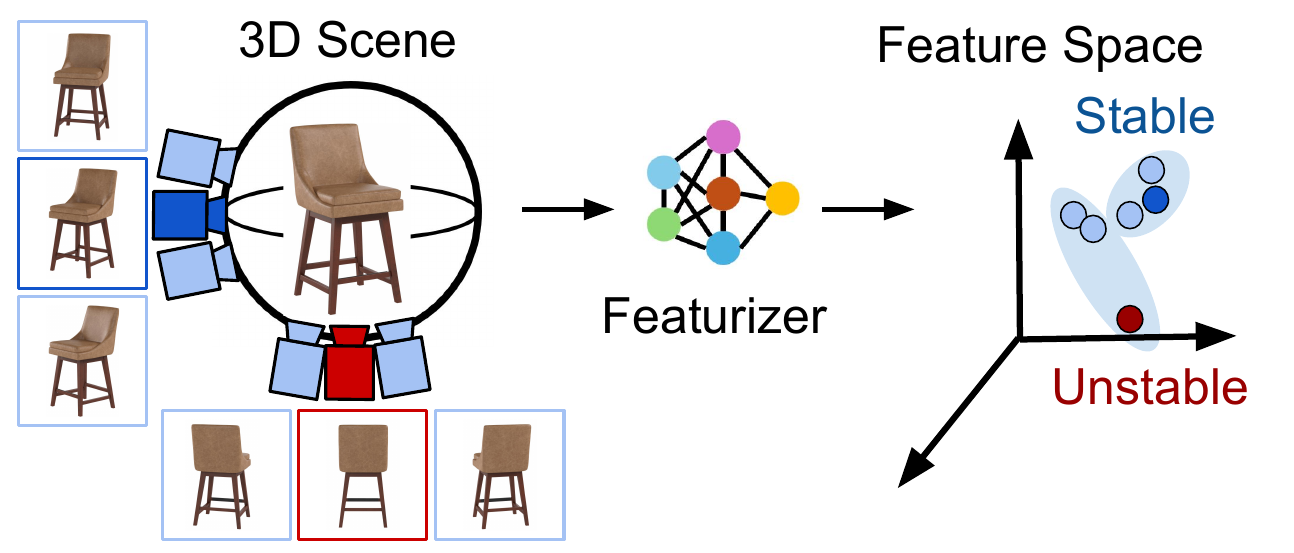}
  \caption{ \textbf{Viewpoint stability of a featurizer.} The blue camera represent small perturbations in camera space that result in correspondingly small changes in a feature space of a featurization function, indicating a stable viewpoint. In contrast, the red camera represent small perturbations in camera space that lead to large changes in feature space, signifying an unstable viewpoint.}
    \label{fig:teaser}
\end{figure}

In the light of the above discussion, we take a deeper look in this study on how popular vision foundation models are affected by changes to viewpoint. In particular, we propose studying their viewpoint stability, or their rate of change of their output features with respect to changes in viewpoint. We define a model's feature output from a certain viewpoint of a scene as unstable if minor changes in the viewing angle lead to significant change in the returned features (see Fig.~\ref{fig:teaser}). A robust representation should ideally exhibit minimal changes in high-level scene content with varying viewpoints. Viewpoint instability may be caused by several factors. As explored in a recent study named OmniView~\cite{omniview}, vision language models (VLMs) such as CLIP~\cite{radford2021learning} and BLIP~\cite{li2022blip} struggle with out-of-distribution (OOD) viewpoints: object viewpoints that are rare in their training distributions. For instance, a mattress could be erroneously perceived as a closed laptop from a frontal viewpoint due to the abundance of closed laptops compared to mattresses in the training data from that particular viewpoint (see Fig. \ref{fig:llava_qual}). 

However, instability is unavoidable in some cases, such as with accidental viewpoints, where a specific camera orientation can obscure the true 3D structure of an object~\cite{accid,accidental_viewpoints,freeman1994generic}. In these cases, there is a fundamental geometric ambiguity that is not explicitly conveyed by featurizers. Understanding when a feature representation is unstable, and whether it coincides with known accidental viewpoints, can improve our grasp of deep learning models' limitations, suggest pathways for enhancing their robustness, and use them in more responsible ways for downstream applications. Existing works with foundation models such as OmniView notably miss critical discussions on accidental viewpoints - a topic generally neglected since the foundational studies of the late 20th century~\cite{accid}. 

In this study, we introduce a methodology for recognizing viewpoint instabilities directly from the output feature representations of foundational models, without the need to access the input images. This capability is particularly beneficial to model featurizer-specific instabilities, and to support privacy-sensitive applications where image access is restricted. We also propose a method to further classify unstable viewpoints into accidental versus other varieties from their features. Through extensive experiments on downstream tasks such as classification, VQA, and monocular 3D reconstruction, we illustrate the influence of viewpoint stability on model performance and identify specific conditions under which model robustness may falter due to viewpoint-related biases.  Our findings first reveal that foundation models, regardless of their structural differences, exhibit a notable consistency in encoding accidental unstable viewpoints. However, their interpretations of OOD viewpoints are variable, which we attribute to inherent model biases. This underscores the need for robust feature representations to support stable performance across diverse viewing conditions.



\section{Related Works}
\label{sec:related}
Viewpoint-invariant reasoning is a fundamental challenge in computer vision. Classical algorithms to address viewpoint variability depend on explicit geometric constraints or multiple view representations in the image space~\cite{biederman1987recognition, freeman1994generic, freeman1996exploiting,zemel1990discovering}. Of particular relevance to our study are accidental viewpoints~\cite{accid,freeman1994generic,accidental_viewpoints}, or unstable viewpoints caused by the camera orientation obscuring an object's true 3D structure~\cite{accid}, which present significant barriers to robust recognition. In contrast to classical approaches, this work forms the first comprehensive study of modern foundation models for viewpoint instability, rather than from images cues directly.

\subsection{Viewpoint Biases in Neural Networks}
Deep neural network classifiers have been known to have viewpoint biases, i.e., performance gaps depending on viewpoint. Popular CNNs used for classification tasks exhibit a noticeable drop in performance for images taken from underrepresented viewpoints~\cite{alcorn2019strike,madan2022and}. Other studies~\cite{zhao2022ood, dong2022viewfool} investigated the biases of models towards canonical viewpoints and introduced datasets and techniques for evaluating viewpoint robustness, focusing on \uncommon (but not explicitly accidental) viewpoints. Specifically, ViewFool~\cite{dong2022viewfool} proposed a technique to generate adversarial viewpoints that push the models towards poor performance, effectively revealing their weaknesses in handling OOD viewpoints. Another study demonstrates predicting OOD viewpoint instabilities of deep networks and achieving viewpoint-invariant visual recognition through adversarial training~\cite{ruan2023towards}. Several other studies also propose network training paradigms~\cite{chen2019cnn, moon2018view,haque2016towards,shang2022learning,wu2014viewpoint} to encourage viewpoint-agnostic learning representations for specific applications.

Though inspired by these studies, our work differs in several ways. First, we focus our attention on a range of modern foundation models rather than traditional classifiers. Second, we aim to predict viewpoint instability directly from features outputted by networks, which offers potential means to flag uncertain responses from these algorithms. Third, we provide a distinct emphasis on understanding how networks react to accidental viewpoints.

\subsection{Robustness Analysis of Foundation Models}
Foundation models, such as CLIP~\cite{radford2021learning}, DINOv2~\cite{oquab2023dinov2}, and Llama~\cite{touvron2023llama}, are now widely used in computer vision due to their state-of-the-art generalization capabilities. As a result, their robustness has also been under scrutiny in a range of contexts, including adversarial attacks~\cite{schlarmann2023adversarial}, dataset content~\cite{nguyen2022quality}, compositional reasoning~\cite{doveh2023dense}, and visual anomalies~\cite{tu2024closer}. Several recent works analyze foundation model features with respect to 3D awareness~\cite{el2024probing,schlarmann2024robust, mao2022understanding}, and Omniview~\cite{omniview} found that the robustness of vision-language models (VLMs) varies significantly based on viewpoints in their training data.

While Omniview is the closest in spirit to our analysis, they did not explicitly study unstable viewpoints (and the further exploration of accidental cases), and only evaluated two foundation models (CLIP and BLIP). In contrast, we systematically evaluate nine foundation models, including CLIP \cite{radford2021learning}, ConvNeXt \cite{liu2022convnet}, Deit III \cite{touvron2022deit}, DINO \cite{caron2021emerging}, DINOv2 \cite{oquab2023dinov2}, DreamSim \cite{fu2023dreamsim}, MAE \cite{he2022masked}, SAM \cite{foret2020sharpness, chen2021vision}, and SigLip \cite{zhai2023sigmoid}. Therefore, our work highlights fundamental gaps in robustness that have not been addressed in prior literature.

\begin{figure*}
    \centering
    \includegraphics[width=\textwidth]{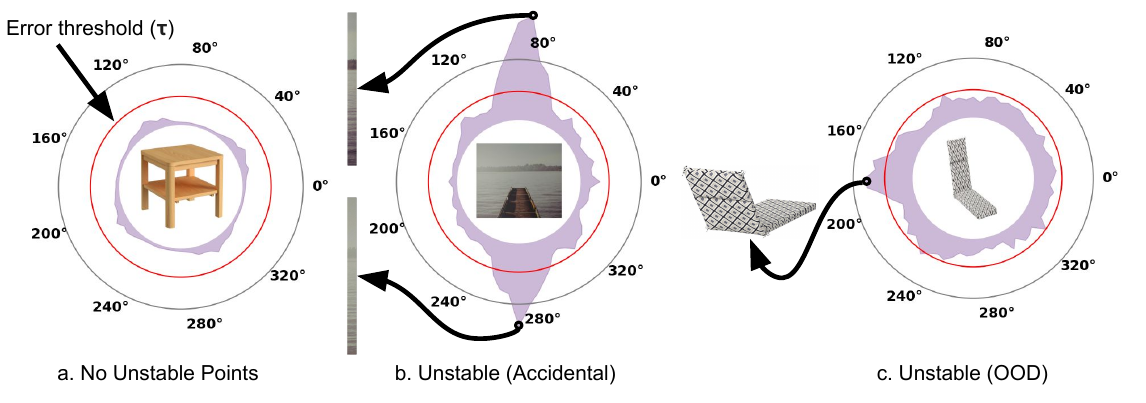}
    \caption{\textbf{Instability scores of CLIP~\cite{radford2021learning} features with respect to viewpoints for three scenes in the ABO~\cite{collins2022abo} dataset.} (Left) a wooden side table with no instabilities. (Center) a painting with two accidental viewpoints, where the structure of the painting is hidden due to viewpoint. (Right) High-back cushion with an \uncommon (OOD) viewpoint.}
    \label{fig:instability_plot_examples}
\end{figure*}
\section{Experimental Analysis}

We present various analyses of foundation models with respect to viewpoint stability in this section. First, we describe the data and models we use in Sec.~\ref{sec:datasets}. Second, we describe our method for quantifying viewpoint instability types in Sec.~\ref{sec:instability-types}. Third, we explore predicting viewpoint instabilities from featurizer outputs alone in Sec.~\ref{sec:predict}. Finally, in Sec.~\ref{sec:downstream} we demonstrate that viewpoint instabilities have significant consequences in various downstream applications using these foundation models.

\subsection{Data and Featurizers}
\subsubsection{Datasets}
\label{sec:datasets}
We used two datasets for our analysis. The first is \textbf{Amazon-Berkeley Objects (ABO)}~\cite{collins2022abo}, consisting of product catalog images, metadata, and artist-created 3D models of real household objects. We specifically used the ``Spins'' subset of ABO, consisting of 8,200 objects distributed across 103 categories, e.g., sofas, wall art, pillows, dressers, and lamps. Each object has 72 image renderings captured at 5-degree azimuth intervals with 360-degree coverage, with randomized elevation angles and varying lighting conditions. In total, this yields 590,400 unique viewpoints across all objects with which we conduct our analysis. Our second dataset is \textbf{Common Objects in 3D (CO3D)} ~\cite {reizenstein2021common}, consisting of 1.5 million real-world multi-view images for roughly 19,000 objects coming from 50 MS COCO~\cite{lin2014microsoft} common object categories. The images are collected by Amazon Mechanical Turk workers, who are instructed to take a 360-degree video of an object from a specified category. 

\subsubsection{Featurizers}\label{sec:featurizers}
We conduct our analysis using nine different deep network featurizers with different training data, training paradigms, and application areas:

\textbf{1. CLIP}~\cite{radford2021learning}: trained using contrastive learning on 400 million (image, text) pairs scraped from the Internet. We used the default public ViT-B/32 architecture. \noindent \textbf{2. DINO}~\cite{caron2021emerging}: self-supervised framework based on student-teacher knowledge distillation, pretrained on ImageNet (without labels)~\cite{russakovsky2015imagenet}.
We used the default public ViT variant with 85.8M parameters. \noindent \textbf{3. DINOv2}~\cite{oquab2023dinov2}: an improved version of DINO that is trained on a curated dataset LVD-142M~\cite{oquab2023dinov2} with 142M images. We used the public ViT variant with 88.6M parameters. \noindent \textbf{4. ConvNeXt}~\cite{liu2022convnet}: CNN family achieving comparable performance to vision transformers. 
We used a public model fine-tuned on ImageNet-1k~\cite{deng2009imagenet} with 846.5M parameters. \noindent \textbf{5. Deit III}~\cite{touvron2022deit}: an approach for training vision transformers that includes techniques such as using 3-Augment for data augmentation, simple random cropping, and low resolutions. We used the image classification variant with 86.6M paramters trained on Imagenet-1k~\cite{deng2009imagenet}. \noindent \textbf{6. DreamSim}~\cite{fu2023dreamsim}: used to compute perceptual similarity between a pair of images, trained on the NIGHTS~\cite{fu2023dreamsim} dataset with 60k images with human perceptual evaluations. We used the standard configuration with ViT-B/16 backbone and 266 million parameters. \noindent \textbf{7. Masked Autoencoder (MAE)}~\cite{he2022masked}: self-supervised model that reconstructs images from partially masked inputs, pre-trained on ImageNet-1K. We used the ViT variant with 630.8 million parameters. \noindent \textbf{8. Sharpness-Aware Minimization (SAM)}~\cite{foret2020sharpness, chen2021vision}: an optimization method that improves generalization and performance of large vision transformers. We used a ViT model with 88.2 million parameters.
parameters. \noindent \textbf{9. SigLip}~\cite{zhai2023sigmoid}: CLIP-based image-text model utilizing sigmoid loss, pre-trained on the WebLI dataset~\cite{chen2022pali}, offering enhanced 0-shot accuracy. We used the public model with 878 million parameters.

\begin{figure*}[t!]
    \centering
    \includegraphics[width=\linewidth]{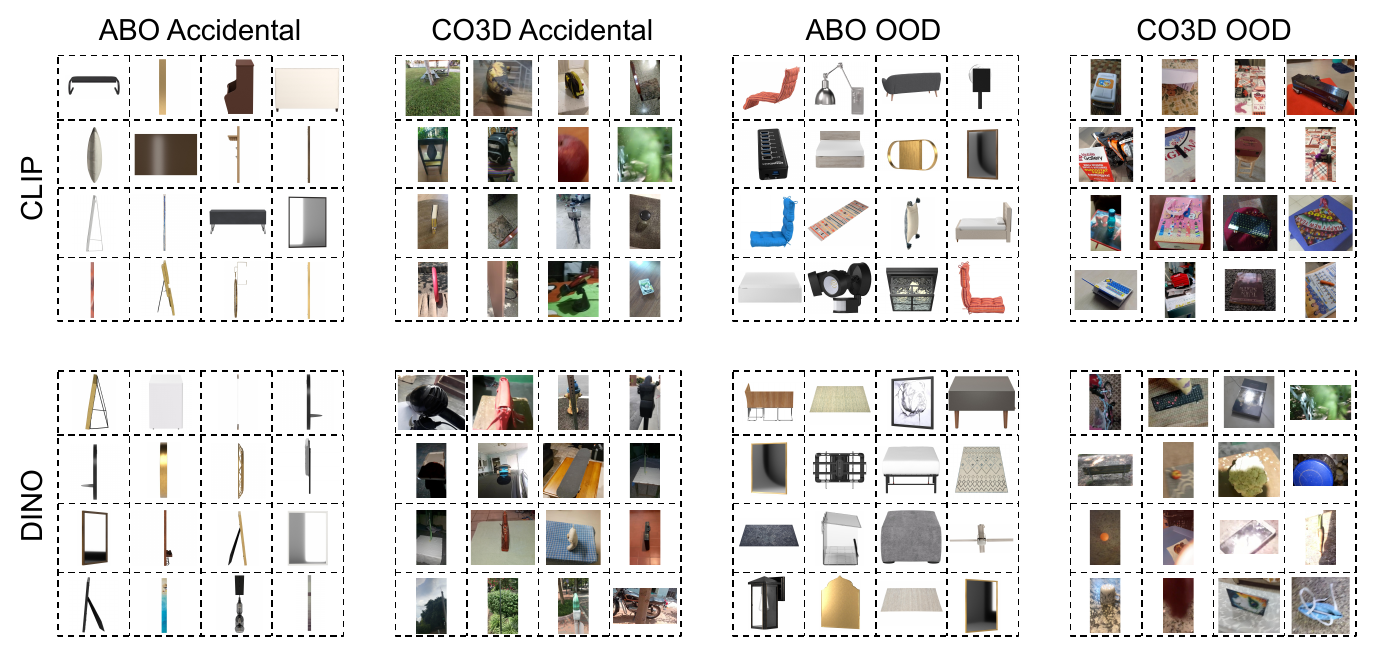} 
\caption{\textbf{Examples of accidental and OOD viewpoints using CLIP and DINO embeddings across the ABO and CO3D datasets.} Accidental views obscure object's true structure, while OOD views present uncommon orientations rarely or never seen during training. In the CO3D dataset, additional sources of instability include occlusions, image blur, and objects displayed upside down. }
    \label{fig:cluster_sample_combined} 
\end{figure*}

\subsection{Viewpoint Instabilities in Foundation Models and Their Types}
\label{sec:instability-types}
Given an image $I_{v} \in \mathbb{R}^{H \times W}$ of a 3D scene taken at view $v$ (3D position and angle), and an arbitrary model \( f(\cdot): \mathbb{R}^{H\times W} \rightarrow \mathbb{R}^d \) that embeds the image into a $d$-dimensional vector, we aim to measure the model's viewpoint stability at $v$, i.e., how the model's output features $f(I_v)$ respond to a small perturbation in $v$. If such a perturbation causes a large change, we consider these features unstable. 

We define two distance functions: $d_v (\cdot, \cdot)$ which measures the distance between two orientations $v_1$ and $v_2$, and $d_f (\cdot, \cdot)$ which measures the distance between two feature vectors $f(v_1)$ and $f(v_2)$. In our experiments, we let $d_v$ encode Euclidean distance based on 3D position and angle, and let $d_f$ encode cosine distance (the most commonly used distance for features of modern foundation models). Further, let $N_v(v_i, r)$ denote the set of all nearest neighbors to $v_i$ such that if $u \in N(v_i, \tau)$, $d_v(v_i, u) \leq r$. We then define the instability score at $v_i$:
\begin{align}
\text{ins}_f(v_i) = \frac{1}{|N_v(v_i, r)|} \sum_{u \in N(v_i, r)} d_f(f(v_i), f(u)).
\label{eq:instability}
\end{align}
We then label \( v_i \) as unstable if $\text{ins}_f(v_i) > \tau$, for some fixed threshold \( \tau \).  

Fig.~\ref{fig:instability_plot_examples} illustrates stability scores and unstable viewpoints detected for three examples from the Amazon-Berkeley Objects (ABO) dataset \cite{collins2022abo}. Some objects, such as the side table (Fig.~\ref{fig:instability_plot_examples}-left), have no unstable viewpoints. Others, such as the painting (Fig.~\ref{fig:instability_plot_examples}-center) have accidental viewpoints corresponding to the camera perfectly aligned with the object to conceal its geometry. And yet other objects, such as the cushion (Fig.~\ref{fig:instability_plot_examples}-right) have instabilities which do not appear accidental, but rather are caused by other biases or artifacts, which we call OOD.  

\subsection{Predicting Viewpoint Instabilities from Features}
\label{sec:predict}
\noindent \textbf{Classifying Stable and Unstable Points.} In most real applications, we will not have access to a dense set of neighboring images to a given view. In these cases, we will be unable to apply Eq.~\ref{eq:instability} to determine whether $f(v_i)$ are stable features or not. Hence, we explore predicting instability from $f(v_i)$ directly. To do so, we trained SVM binary classifiers~\cite{cortes1995support} with RBF kernels for all featurizers on both datasets to predict view instability, using Eq.~\ref{eq:instability} to produce ground truth binary labels. For each (featurizer,dataset) combination, we set the threshold $\tau$ to the 97th percentile of instability scores computed across all viewpoints. We split the datasets into 80/20 train/test splits along scenes, such that images from the same scene do not span both groups. To reduce redundant examples, we applied non-maximal suppression to only choose maximally unstable points in local neighborhoods. Table~\ref{tab:stage_comparison_abo} presents classification results. We obtained accuracies of [81-96]\% across featurizers on ABO, and [71-77]\% on CO3D, demonstrating the surprising finding that view stability is largely predictable from features alone. MAE is generally the hardest feature space in which to separate its stable and unstable views. \\

\begin{figure}[t!]
    \centering
    \includegraphics[width=\linewidth, trim={0cm 0cm 0 0}, clip]{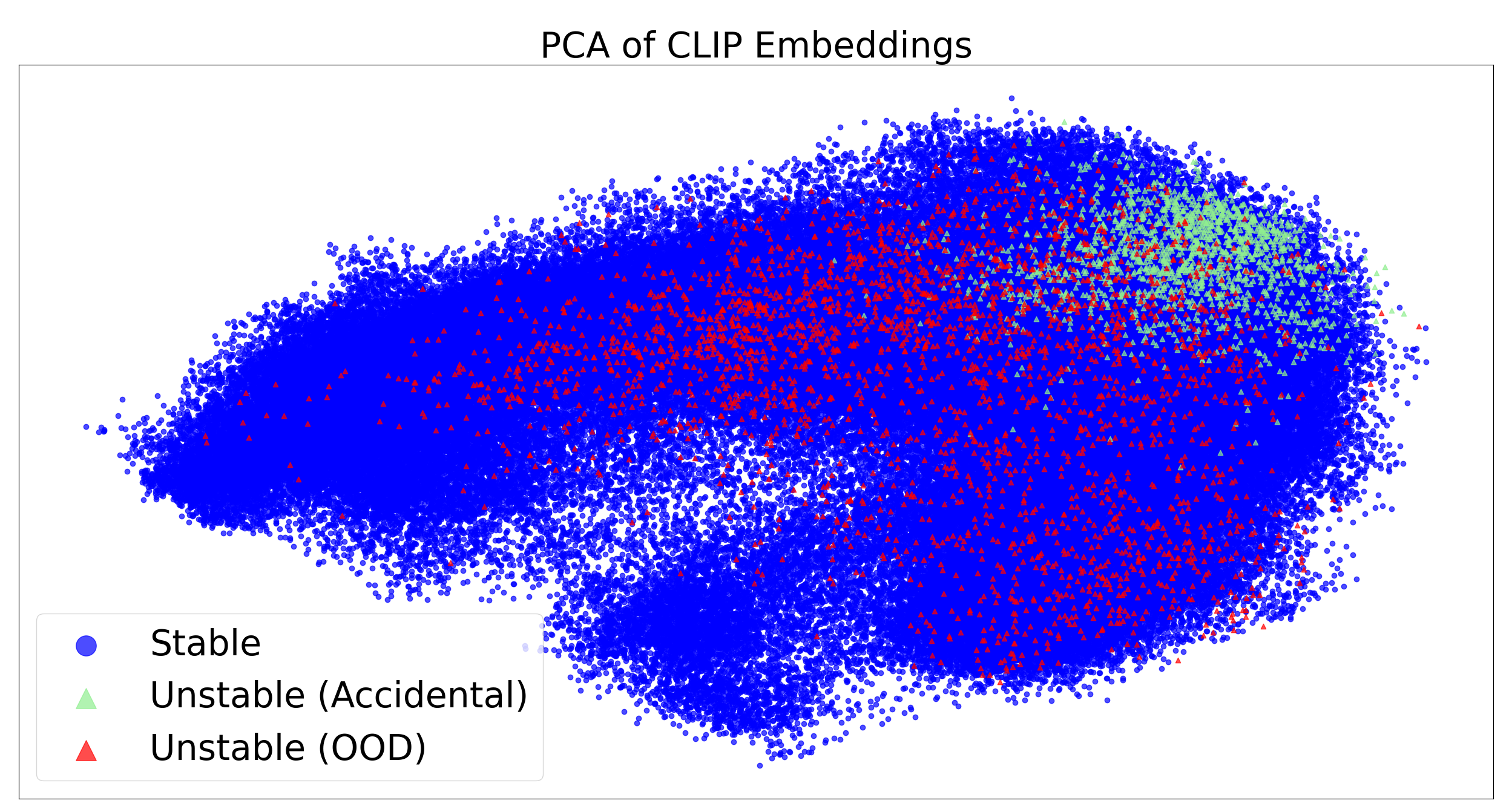} 
    \caption{ \textbf{PCA visualization of stable (blue), unstable-accidental (green), and unstable-OOD (red) viewpoints using CLIP embeddings from ABO dataset} We see that unstable viewpoints generally cluster in feature space, with accidental viewpoints tightly grouped together. Please refer to sample images from the unstable-accidental and unstable-OOD in Fig.~\ref{fig:cluster_sample_combined}. } 
    \label{fig:pca_abo_clip} 
\end{figure}

\begin{table}[t!]
    \centering
    \caption{\textbf{Classification accuracy for separating stable and unstable viewpoints in various feature spaces.} We obtain stable/unstable labels per viewpoint using Eq.~\ref{eq:instability}. We obtain surprisingly high classification scores across all featurizers, with better performance on ABO due to less variabilities (backgrounds, lighting, artifacts).}
    \begin{tabular}{|c|c|c|}
        \hline
        \textbf{Model} & ABO & CO3D \\
        \hline
        CLIP & 88.50\% & 72.37\% \\
        ConvNeXt & 90.73\% & 76.18\% \\
        DeiT III & 88.05\% & 76.85\% \\
        DINO & 94.08\% & 77.08\% \\
        DINOv2 & 92.79\% & 73.78\% \\
        DreamSim & 96.60\% & 73.37\% \\
        MAE & 81.68\% & 71.52\% \\
        SAM & 89.53\% & 72.33\% \\
        SigLip & 87.97\% & 74.65\% \\
        \hline
    \end{tabular}
    \label{tab:stage_comparison_abo}
\end{table}

\noindent \textbf{Predicting Accidental and OOD Unstable Subtypes.} We further hypothesized that unstable view subtypes of accidental and OOD may also be predictable from features alone. Fig.~\ref{fig:pca_abo_clip} shows a 2D projection of CLIP features on the ABO dataset using Principal Component Analysis (PCA), with stable points displayed in blue, and two clusters of unstable points obtained with $k$-means shown in green and red. Upon visual inspection of samples, we find the green cluster largely corresponds to accidental views, while the red are other out-of-distribution (OOD) views. Interestingly, the green cluster is also ``tighter,'' which agrees intuitively with the idea that these views share similar collapsed 1D/2D structures. We therefore explore separating accidental from OOD unstable viewpoints in a feature space by fitting $k=2$ clusters on unstable points, and assigning points that fall in the cluster with highest silhouette score \cite{rousseeuw1987silhouettes} as accidental.

Using this approach, we obtained accidental and OOD clusters for all featurizers, and present random samples from the CLIP and Dino models in Fig.~\ref{fig:cluster_sample_combined}. The CO3D dataset, which includes more complex backgrounds and in-the-wild images, also reveals accidental viewpoints that arise when objects are temporarily occluded. For CO3D OOD views, we often observe objects that appear upside down or temporarily blurry, likely due to changes in focus along the camera’s trajectory. We found similar trends for all other featurizers (see Supplementary).

We next analyzed whether images from the predicted accidental clusters agree with human perception of accidental viewpoints. Using a human annotator, we obtained $1\,000$ true accidental and $1\,000$ non-accidental viewpoints from ABO. We found an average label overlap of 78.3\% of featurizers with these ground truth labels, with DINO having the highest agreement of 88.19\%, and CLIP having the lowest agreement with 68.84\%. When we consider all predicted unstable viewpoints (both accidental and OOD) as accidental, the accuracy with respect to humans increases to 89.8\%, with SigLip achieving the highest agreement at 94.1\% and MAE the lowest at 81.3\%. This indicates that general instability is far easier to predict in feature space compared to unstable subtypes. 


Finally, we assessed the level of agreement across featurizers on identifying accidental and OOD viewpoints. Using Intersection over Union (IoU), a standard metric for quantifying overlap between sets, we measured agreement in viewpoint classification. We observed a high level of agreement for stable viewpoints (mean IoU of 0.99) and moderate agreement for accidental viewpoints (0.52), indicating that these viewpoints are generally identified similarly across featurizers. However, agreement on OOD viewpoints was markedly low, with a mean IoU of only 0.12. We provide pairwise IoU heatmaps for each label category (stable, accidental, and OOD) across all 9 featurizers in Supplementary.


\subsection{Downstream Applications}
\label{sec:downstream}
The above analyses highlight viewpoint instabilities in the feature spaces of popular foundation models. To understand the practical impact of these instabilities on real-world applications, we next investigated three downstream tasks: classification (zero-shot and linear probing), visual question answering (VQA), and monocular 3D reconstruction.

\subsubsection{Zero-Shot Classification}
In this experiment, we evaluated CLIP's zero-shot classification performance on both ABO and CO3D datasets. Leveraging CLIP’s shared feature space, we encode ground truth (GT) category labels as text embeddings and compute feature embeddings for each image, assigning the label with the minimum cosine distance to each image embedding.
We report classification performance using accuracy@k, which reflects the percentage of cases where the GT label is among the top \( k \) closest predicted labels, with \( k \) chosen from \{1, 5, 10\}.
The results in Table~\ref{tab:clip_0shot_combined} show a sharp drop in accuracy for unstable viewpoints relative to stable ones, particularly for OOD, and even more so for accidental. This illustrates the model's challenge in maintaining reliable classification under less common or ambiguous viewing conditions.
\begin{table}[t!]
\centering
\caption{\textbf{Zero-shot classification accuracy of CLIP for stable and unstable viewpoints at different Top-K values on the ABO and CO3D datasets.} For the ABO dataset, Acc@5 is marked in bold as the most suitable measure due to overlapping object categories, while for the CO3D dataset, Acc@1 is marked in bold given the distinct categories. The model shows a decline in performance for OOD and accidental viewpoints across both datasets.}
\begin{tabular}{|c|c|c|c|c|}
\hline
\multicolumn{4}{|c|}{\textbf{ABO Dataset} } \\ \hline
&  Stable                     & OOD & Accidental \\ \hline
Acc@1  & 40.23 & 22.47 & 0.84 \\
\textbf{Acc@5}  & 68.74 & 42.56 & 3.00 \\
Acc@10 & 79.97 & 56.05 & 4.85 \\
\hline
\multicolumn{4}{|c|}{\textbf{CO3D Dataset}} \\ \hline
&  Stable                     & OOD & Accidental \\ \hline
\textbf{Acc@1}  & 78.58  & 68.45 & 48.36 \\
Acc@5  & 93.77 & 90.45 & 75.49  \\
\hline
\end{tabular}
\label{tab:clip_0shot_combined}
\end{table}
\subsubsection{Linear Probe Classification}
While CLIP’s shared image-text space allows for zero-shot classification, most other foundation models lack this capability. Linear probing -- or simply training an extra linear prediction layer using features outputted by a backbone network -- is a common strategy used to evaluate the usefulness of a generic feature space. We trained a single linear layer on top of each model’s embeddings for 100 epochs, using the Adam optimizer~\cite{diederik2014adam}, and a learning rate of 1e-3. We applied an 80-20 train-test split, ensuring no overlap of objects between the sets to avoid data leakage. The results, shown in Table~\ref{tab:linear_probe_combined}, indicate a substantial accuracy decline for unstable viewpoints, underscoring the limitations of current models in handling viewpoint variability and highlighting the need for more viewpoint-robust representations.

\begin{table}[t!]
\centering
\caption{\textbf{Classification accuracy using linear probe for stable and unstable viewpoints on the ABO and CO3D datasets.} The results show a significant decline in accuracy for both OOD and accidental viewpoints across models.}
\begin{tabular}{|c|c|c|c|}
\hline
\multicolumn{4}{|c|}{\textbf{ABO Dataset}} \\ \hline
& Stable & OOD & Accidental \\ \hline
\textbf{CLIP}       & 89.62\% & 73.97\% & 63.19\% \\
\textbf{ConvNeXt}   & 99.08\% & 88.35\% & 79.29\% \\
\textbf{Deit III}   & 91.43\% & 73.00\% & 63.80\% \\
\textbf{DINO}       & 92.37\% & 78.41\% & 69.35\% \\
\textbf{DINOv2}     & 95.26\% & 76.39\% & 72.18\% \\
\textbf{DreamSim}   & 97.96\% & 86.93\% & 79.89\% \\
\textbf{MAE}        & 93.97\% & 89.72\% & 70.93\% \\
\textbf{SAM}        & 91.79\% & 75.49\% & 63.77\% \\
\textbf{SigLip}     & 95.95\% & 81.82\% & 73.77\% \\
\hline
\multicolumn{4}{|c|}{\textbf{CO3D Dataset}} \\ \hline
& Stable & OOD & Accidental \\ \hline
\textbf{CLIP}       & 98.84\% & 95.22\% & 77.46\% \\
\textbf{ConvNeXt}   & 99.74\% & 96.51\% & 87.34\% \\
\textbf{Deit III}   & 96.08\% & 74.57\% & 73.45\% \\
\textbf{DINO}       & 96.71\% & 83.41\% & 70.19\% \\
\textbf{DINOv2}     & 98.94\% & 84.01\% & 77.49\% \\
\textbf{DreamSim}   & 98.75\% & 93.12\% & 76.96\% \\
\textbf{MAE}        & 93.97\% & 90.26\% & 66.49\% \\
\textbf{SAM}        & 93.92\% & 73.81\% & 64.12\% \\
\textbf{SigLip}     & 99.85\% & 99.10\% & 81.21\% \\
\hline
\end{tabular}
\label{tab:linear_probe_combined}
\end{table}

\begin{figure}[t!]
\centering
\includegraphics[width=\linewidth, trim={8cm 2cm 0 0}, clip]{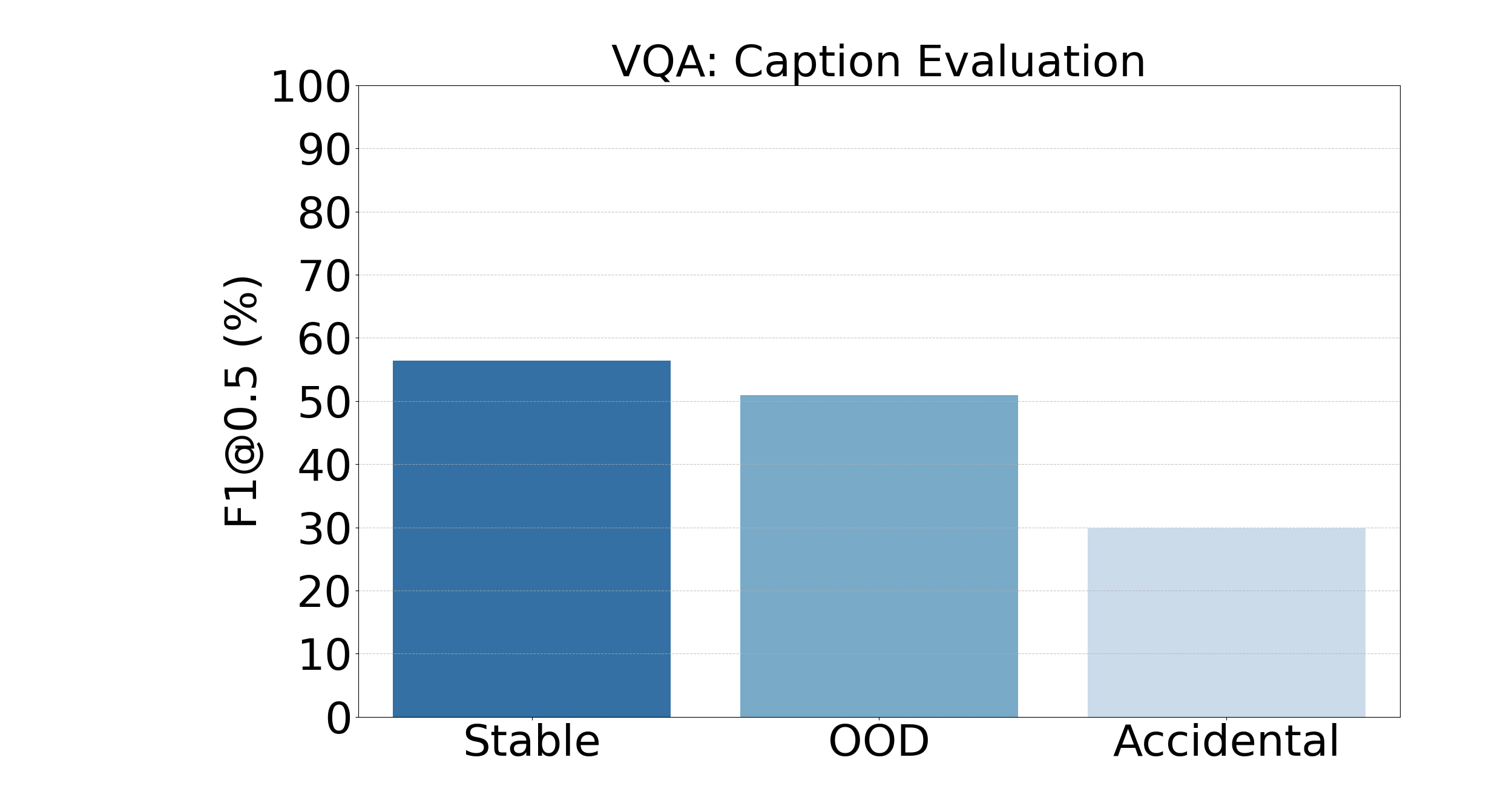}
\caption{\textbf{Percentage of generated captions achieving an F1 score above 0.5 (F1@0.5) (higher is better) for stable, OOD, and accidental viewpoints on the ABO dataset.} The F1 score is calculated using BERTScore between generated captions and ground truth (GT) captions, where GT captions are generated with access to the object's label information. Captions generated from stable viewpoints achieve higher F1 scores compared to those from accidental viewpoints, indicating that viewpoint instability affects caption generation accuracy.}
\label{fig:vqa_abo_quant}
\end{figure}

\begin{figure}[t!]
\centering
\includegraphics[width=\linewidth, trim={8cm 2cm 0 0}, clip]{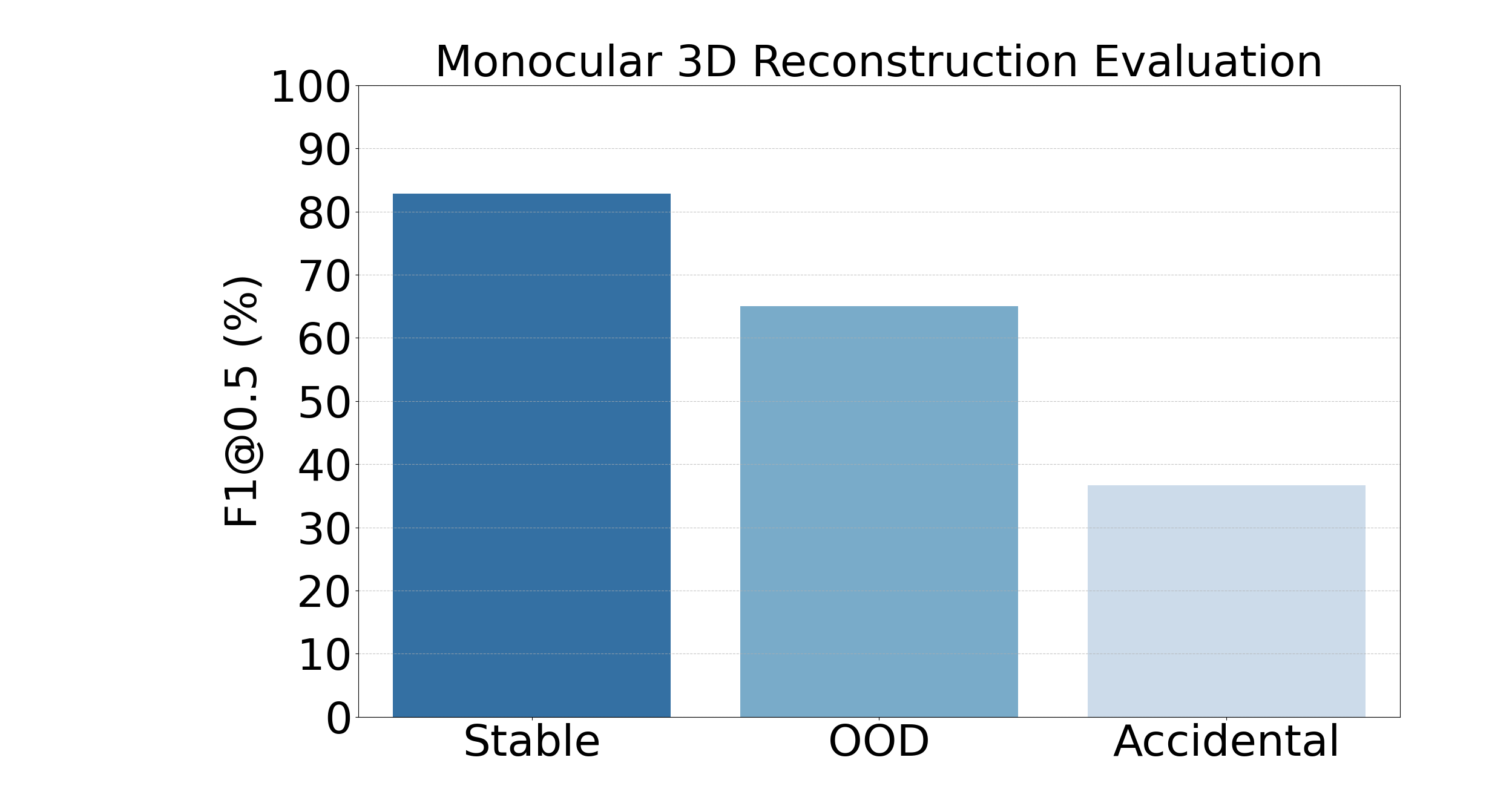}
\caption{\textbf{Percentage of reconstructed 3D models achieving an F1 score above 0.5 (F1@0.5) (higher is better) for stable, OOD, and accidental viewpoints on the ABO dataset.} The F1 score, calculated between reconstructed models and ground truth (GT) models, indicates that reconstructions from stable viewpoints outperform those from OOD and accidental viewpoints.}
\label{fig:3dr_abo_quant}
\end{figure}

\begin{figure*}[t!]
    \centering
    \includegraphics[width=\textwidth]{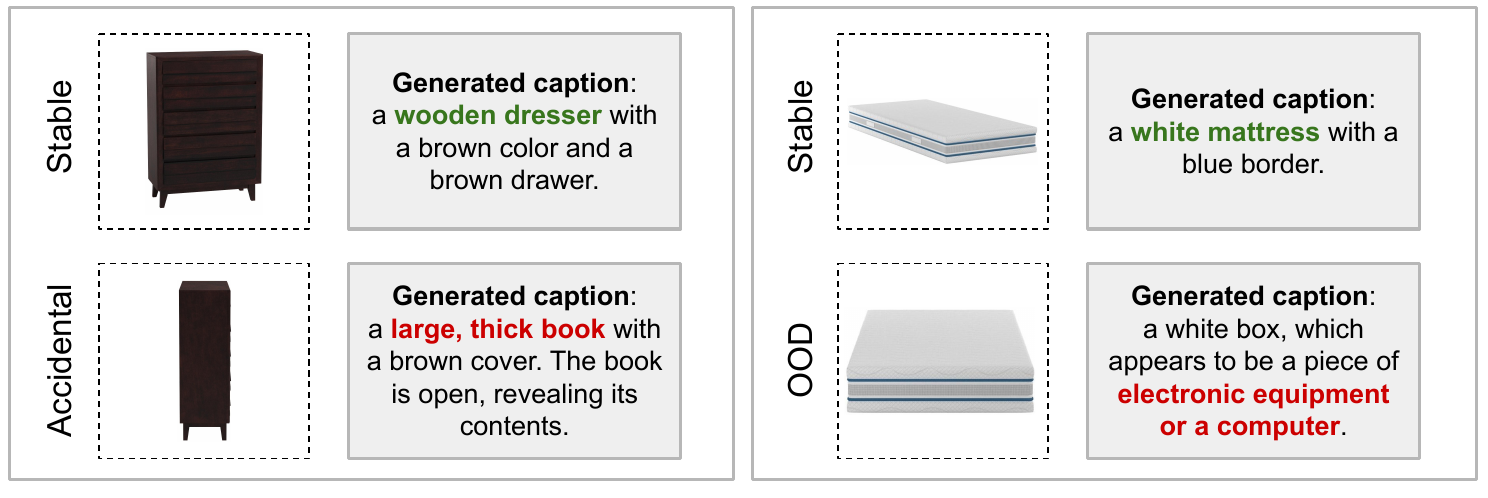}
\caption{\textbf{Examples of generated captions for stable, OOD, and accidental viewpoints using LLaVA-1.5  \cite{liu2024improved} (CLIP \cite{radford2021learning} backbone).} Captions for stable viewpoints are factual, while those for accidental and OOD viewpoints often contain inaccuracies. Here, an accidental viewpoint causes LlaVA to describe a dresser as a book, and an OOD viewpoint causes a misinterpretation of a mattress as a computer.}
\label{fig:llava_qual}
\end{figure*}

\begin{figure*}[t!]
    \centering
    \includegraphics[width=\linewidth]{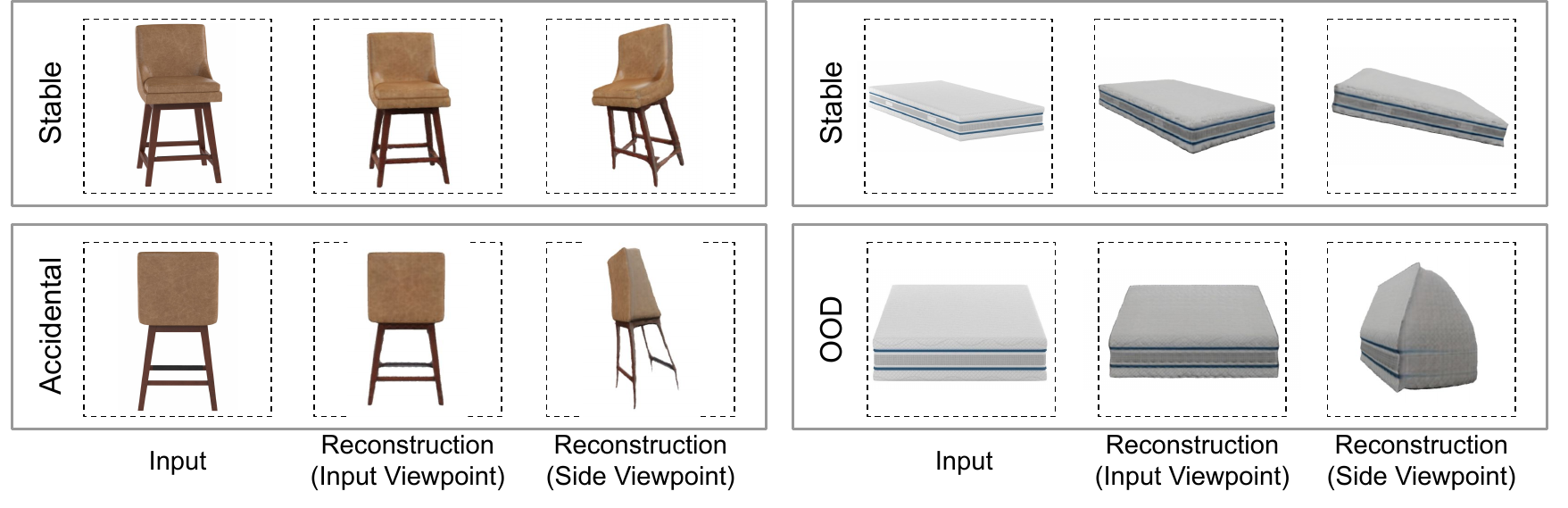}
\caption{\textbf{Single-view reconstruction results for stable and unstable viewpoints from the ABO dataset, using Stable Fast 3D~\cite{boss2024sf3d} with DINOv2~\cite{oquab2023dinov2} as the image featurizer.} Stable viewpoints yield reconstructions with accurate geometry, while accidental and OOD viewpoints lead to distortions.}
    \label{fig:3dr_combined}
\end{figure*}

\subsubsection{Visual Question Answering (VQA)}\label{sec:vqa}
We next conducted a VQA experiment using LLaVA-1.5~\cite{liu2024improved}, which leverages CLIP as its backbone, to evaluate how well it generates descriptive captions for images across stable, accidental, and OOD viewpoints. Prompted with a generic question, \textit{What is the content of the image?}, we assessed the model’s accuracy in describing the content under different viewing conditions. Qualitative examples in Fig.~\ref{fig:llava_qual} show that for stable viewpoints, LLaVA-1.5 generates accurate captions, whereas for accidental and OOD viewpoints, the model often fails, misidentifying objects or adding irrelevant details. To quantify performance, we compared generated captions to ground truth (GT) captions. GT captions were created using LLaVA-1.5 with an informative prompt, \textit{What's the content of the image knowing that the main object is \{label\}}, where \{label\} represents the known category from the ABO dataset. 
We measured the percentage of generated captions with an F1 score above 0.5 (F1@0.5), which we consider indicative of acceptable alignment with the ground truth. The F1 score, computed as the harmonic mean of precision and recall, was calculated using BERTScore~\cite{zhang2019bertscore}. Results in Fig.~\ref{fig:vqa_abo_quant} show significantly higher F1@0.5 for stable viewpoints compared to accidental, underscoring the challenges that viewpoint instability poses for accurate content description.

\subsubsection{Monocular 3D Reconstruction}\label{sec:m3dr}
To examine the effects of viewpoint instability on 3D reconstruction, we conducted experiments using Stable Fast3D \cite{boss2024sf3d}, a state-of-the-art monocular 3D reconstruction model that uses DINOv2 as its image featurizer. 
We generated single-view reconstructions for stable, accidental, and OOD viewpoints. After reconstruction, we aligned the predicted shapes with ground truth (GT) models using RANSAC \cite{fischler1981random} followed by ICP \cite{chen1992object} to ensure optimal alignment.
Figures~\ref{fig:3dr_combined} illustrate the differences in reconstruction quality across viewpoint types. Stable viewpoints yield accurate reconstructions with well-preserved geometry. 

In contrast, accidental viewpoints—where depth perception is almost absent due to insufficient depth cues—cause the model to incorrectly infer the 3D shape, resulting in collapsed or distorted reconstructions. OOD viewpoints similarly degrade performance, as the atypical angles can cause the model to interpret the object as belonging to an entirely different category, leading to substantial inaccuracies in the reconstruction.
Quantitative results in Fig.~\ref{fig:3dr_abo_quant} confirm these observations, with stable viewpoints achieving higher F-scores \cite{knapitsch2017tanks} relative to OOD and accidental viewpoints. These findings underscore the impact of viewpoint instability on reconstruction accuracy and highlight the need for models capable of adapting to challenging and atypical viewing conditions.

\section{Discussion and Conclusion}
\label{sec:discussion}
In this study, we investigated the viewpoint stability of popular vision foundation models and introduced methodologies for recognizing and classifying instabilities directly from their feature representations. Our experiments first demonstrate that unstable viewpoints are to some degree separable from stable viewpoints in the feature spaces of all the studied foundation models. Compared to the findings past studies demonstrating that styles~\cite{patashnik2021styleclip} and even 3D concepts~\cite{el2024probing} are separable in these feature spaces, our results are surprising because stability is not typically related to semantic learning.

All models exhibited some level of view instability. We found that CLIP, possibly the most popular of all the foundation models we studied, was less stable compared to alternatives like ConvNeXt and DreamSim, particularly in relation to accidental viewpoints. Interestingly, DreamSim uses an ensemble of CLIP, OpenCLIP, and DINOv2, which may explain its relative stability compared to the other algorithms. Foundation models generally agreed on accidental viewpoints, but varied on what they considered OOD. This is reasonable since accidental views are directly defined by the scene geometry and camera angle, while OOD views are also affected by the datasets, architectures, and training paradigms of each network. 

Our analyses also demonstrate that viewpoint instabilities of featurizers lead to tangible dropoffs in performance of downstream models that use them for various applications. Example tasks we explored include classification variants, VQA, and 3D reconstruction. We used two datasets in our analyses: ABO and CO3D. Due to its synthetic nature, ABO yielded far cleaner separation of stable and unstable points in feature spaces. In contrast, due to backgrounds, occlusions, lighting variations, and camera acquisition factors, CO3D yielded less clear separability. And yet, even in these natural conditions, we observed significant adverse effects on downstream tasks for detected unstable views. 

These analyses raise important questions on how to prevent the unsafe use of foundation models with respect to viewpoint instabilities. One natural follow-up to the results of this study is to develop foundation models that output stability confidence measures along with their features. Downstream models can use such confidence measures to signal that its resulting answers are also unreliable, e.g., such as when answering a question for VQA, instead of feigning confidence. Another useful direction to explore is to apply regularization techniques to ensure that small changes in camera space yield correspondingly small changes in feature space. This would approximate the ideal scenario where image embeddings remain independent of camera viewpoint, harking back to classical notions of viewpoint invariance for reasoning in computer vision~\cite{accid, accidental_viewpoints, freeman1994generic}.
\\

\noindent\textbf{Acknowledgement}
Supported by the Intelligence Advanced Research Projects Activity (IARPA) via Department of
Interior/ Interior Business Center (DOI/IBC) contract number 140D0423C0076. The U.S.
Government is authorized to reproduce and distribute reprints for Governmental purposes
notwithstanding any copyright annotation thereon. Disclaimer: The views and conclusions
contained herein are those of the authors and should not be interpreted as necessarily
representing the official policies or endorsements, either expressed or implied, of IARPA,
DOI/IBC, or the U.S. Government.




{
    \small
    \bibliographystyle{ieeenat_fullname}
    \bibliography{main}
}


\end{document}